\documentclass[10pt,twocolumn,letterpaper]{article}

\usepackage{cvpr}
\usepackage{times}
\usepackage{epsfig}
\usepackage{graphicx}
\usepackage{amsmath}
\usepackage{amssymb}

\usepackage[pagebackref=false,breaklinks=true,letterpaper=true,colorlinks,bookmarks=false]{hyperref}

\cvprfinalcopy 

\def\cvprPaperID{2777} 

\ifcvprfinal\fi
\graphicspath{{figures/}} 
\begin{document}

\title{Appearance-and-Relation Networks for Video Classification}

\author{Limin Wang\textsuperscript{1,2} \quad \quad Wei Li\textsuperscript{3} \quad \quad Wen Li\textsuperscript{2} \quad \quad Luc Van Gool\textsuperscript{2} \\
\textsuperscript{1}State Key Laboratory for Novel Software Technology, Nanjing University, China \\
\textsuperscript{2}Computer Vision Laboratory, ETH Zurich, Switzerland  \quad \quad \textsuperscript{3}Google Research \\
}
\maketitle

\begin{abstract}
Spatiotemporal feature learning in videos is a fundamental problem in computer vision. This paper presents a new architecture, termed as {\em Appearance-and-Relation Network} (ARTNet), to learn video representation in an end-to-end manner. ARTNets are constructed by stacking multiple generic building blocks, called as {\em SMART}, whose goal is to simultaneously model appearance and relation from RGB input in a separate and explicit manner. Specifically, SMART blocks decouple the spatiotemporal learning module into an appearance branch for spatial modeling and a relation branch for temporal modeling. The appearance branch is implemented based on the {\em linear combination} of pixels or filter responses in each frame, while the relation branch is designed based on the {\em multiplicative interactions} between pixels or filter responses across multiple frames. We perform experiments on three action recognition benchmarks: Kinetics, UCF101, and HMDB51, demonstrating that SMART blocks obtain an evident improvement over 3D convolutions for spatiotemporal feature learning. Under the same training setting, ARTNets achieve superior performance on these three datasets to the existing state-of-the-art methods.~\footnote{The code is at \url{https://github.com/wanglimin/ARTNet}.} 
\end{abstract}

\section{Introduction}
Deep learning has witnessed a series of remarkable successes in computer vision. In particular, Convolutional Neural Networks (CNNs)~\cite{lecun-98} have turned out to be effective for visual tasks in image domain, such as image classification~\cite{KrizhevskySH12,HeZRS16,SimonyanZ14a,SzegedyLJSRAEVR15}, object detection~\cite{GirshickDDM14}, and semantic segmentation~\cite{LongSD15}. Deep models have been also introduced into video domain for action recognition~\cite{CarreiraZ17,Wang0T15,WangFG16,SimonyanZ14,WangXWQLTV16,TranBFTP15,KarpathyTSLSF14,ZhangWW0W16}, and obtain comparable or better recognition accuracy to those traditional methods with hand-crafted representations~\cite{Laptev05,WangS13a,Wang0T16}. However, the progress of architecture design and representation learning in video domain is much slower, partially due to its inherent complexity and higher dimension. Video could be viewed as the temporal evolution of a sequence of static images. It is generally assumed that two visual cues are crucial for video classification and understanding: (1) {\em static appearance} in each frame, and (2) {\em temporal relation} across multiple frames. Therefore, an effective deep architecture should be able to capture both information to achieve excellent classification accuracy.

\begin{figure*}
\includegraphics[width=\linewidth]{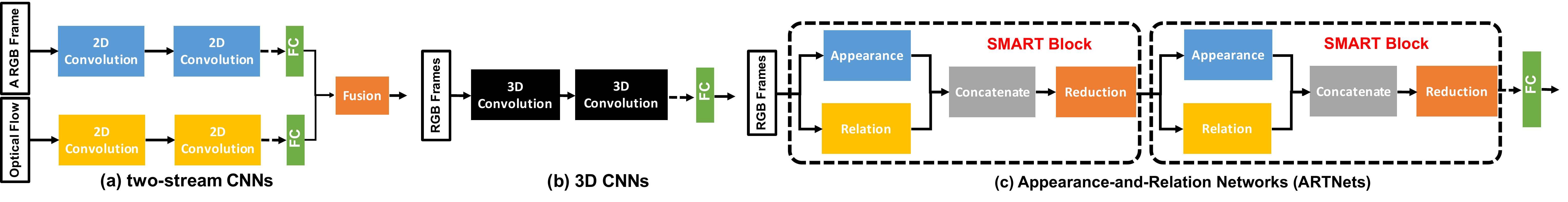}
\caption{{\bf Video architecture comparison}: Our Appearance-and-Relation Networks (ARTNets) are constructed based on the SMART building block, which aims to simultaneously model appearance and relation from RGB in a separate and explicit way. In contrast, two-stream CNNs model them with two inputs and 3D CNNs model them jointly and implicitly with a single 3D convolution.}
\label{fig:net}
\vspace{-4mm}
\end{figure*}

There are three kinds of successful architectures or frameworks for video classification~\cite{CarreiraZ17}: (1) two-stream CNNs~\cite{SimonyanZ14}, (2) 3D CNNs~\cite{JiXYY13,TranBFTP15}, and (3) 2D CNNs with temporal models on top such as LSTM~\cite{DonahueHGRVDS15,NgHVVMT15}, temporal convolution~\cite{NgHVVMT15}, sparse sampling and aggregation~\cite{WangXWQLTV16}, and attention modeling~\cite{WangXLV17,winner17,GanWYYH15}. Two-stream CNNs capture appearance and motion information with different streams, which turn out to be effective for video classification. Yet, it is time-consuming to train two networks and calculate optical flow in advance. To overcome this limitation, 3D CNNs employ 3D convolutions and 3D pooling operations to directly learn spatiotemporal features from stacked RGB volumes. However, the performance of 3D CNNs is still worse than two-stream CNNs, and it is still unclear whether this straightforward 3D extension over 2D convolution could efficiently model static appearance and temporal relation. 2D CNNs with temporal models usually focus on capturing coarser and long-term temporal structure, but lack capacity of representing finer temporal relation in a local spatiotemporal window.

In this paper, we address the problem of capturing appearance and relation in video domain, by proposing a new architecture unit termed as {\em SMART} block. Our SMART block aims to Simultaneously Model Appearance and RelaTion from RGB input in a separate and explicit way with a two-branch unit, in contrast to modeling them with two-stream inputs~\cite{SimonyanZ14} or jointly and implicitly with a 3D convolution~\cite{TranBFTP15}. As shown in Figure~\ref{fig:net}, our SMART block is a multi-branch architecture, which is composed of appearance branch and relation branch, and fuses them with a concatenation and reduction operation. The appearance branch is based on the {\em linear combination} of pixels or filter responses in each frame to model spatial structure, while the relation branch is based on the {\em multiplicative interactions} between pixels or filter responses across multiple frames to capture temporal dynamics. Specifically, the appearance branch is implemented with a standard 2D convolution and the relation branch is implemented with a square-pooling structure. The responses from two branches are further concatenated and reduced to a more compact representation.

A SMART block is a basic and generic building module for video architecture design. For video classification, we present an appearance-and-relation network (ARTNet) by stacking a collection of SMART blocks. Essentially, the appearance and relation information in video domain exhibit multi-scale spatiotemporal structure. The ARTNet is able to capture this visual structure in a hierarchical manner, where SMART units in the early layers focus on describe local structure in a short term, while the ones in the later layers can capture increasingly coarser and longer-range visual structure. An ARTNet is a simple and general architecture which offers flexible implementations. In the current implementation of this paper, the ARTNet is instantiated with the network of C3D-ResNet18~\cite{Tran17} for an engineering compromise between accuracy and computation consumption. Moreover, our ARTNet is complementary to those long-term temporal models, which means any of them could be employed to enhance its modeling capacity. As an example, we use the framework of temporal segment network (TSN)~\cite{WangXWQLTV16} to jointly train ARTNets from a set of sparsely sampled snippets and further improve the recognition accuracy.

We test the ARTNet on the task of action recognition in video classification. Particularly, we first study the performance of the ARTNet on the Kinetics dataset~\cite{KayCSZHVVGBNSZ17}. We observe that our ARTNet obtains an evident improvement over C3D, and superior performance to the exiting state-of-the-art methods on this challenging benchmark under the setting of training from scratch with only RGB input. To further demonstrate the generality of ARTNet, we also transfer its learned video representation to other action recognition benchmarks including HMDB51~\cite{KuehneJGPS11} and UCF101~\cite{Soomro12}, where performance improvement is also achieved.

The main contribution of this paper is three-fold: (1) A SMART block is designed to simultaneously capture appearance and relation in a separate and explicit way. (2) An ARTNet is proposed by stacking multiple SMART blocks to model appearance and relation information from different scales, which also allows for optimizing the parameters of SMART blocks in an end-to-end way. (3) ARTNets are empirically investigated on the large-scale Kinetics benchmark and state-of-the-art performance on this dataset is obtained under the setting of using only RGB input and training from scratch. 

\section{Related Work}

{\bf Deep learning for video classification.} Since the breakthrough of Convolutional Neural Networks (CNN)~\cite{lecun-98} in image classification~\cite{KrizhevskySH12}, several works have tried to design effective architectures for video classification and action recognition~\cite{KarpathyTSLSF14,SimonyanZ14,TranBFTP15,WangXWQLTV16,NgHVVMT15,DonahueHGRVDS15,CarreiraZ17,QiuYM17,SunJYS15}. Karpathy~\emph{et al.}~\cite{KarpathyTSLSF14} first tested deep networks with different temporal fusion strategies on a large-scale and noisily-labeled dataset (Sports-1M) and achieved lower performance than traditional features~\cite{WangS13a}. Simonyan \emph{et al.}~\cite{SimonyanZ14} designed a two-stream architecture containing spatial and temporal nets by explicitly exploiting pre-trained models and optical flow calculation. Tran \emph{et al.}~\cite{TranBFTP15} investigated 3D CNNs~\cite{JiXYY13} on realistic and large-scale video datasets and further studied deep ResNet with 3D convolution~\cite{Tran17}. Carreira~\emph{et al.} proposed a new Two-Stream Inflated 3D CNNs based on 2D CNN inflation, which allows for pre-training with ImageNet models. Meanwhile, several papers~\cite{NgHVVMT15,DonahueHGRVDS15,WangXWQLTV16} tried to model long-term temporal information for action understanding. Ng \emph{et al.}~\cite{NgHVVMT15} and Donahue \emph{et al.}~\cite{DonahueHGRVDS15} utilized the LSTM~\cite{HochreiterS97} to capture the long range dynamics for action recognition. Wang \emph{et al.}~\cite{WangXWQLTV16} designed a  temporal segment network (TSN) to perform sparse sampling and temporal fusion, which aims to learn from the entire video. 

Our work focuses on short-term temporal modeling and is most related with 3D CNNs. Our ARTNet mainly differs to 3D CNNs in that we design a new SMART block to model appearance and relation separately and explicitly with a two-branch architecture, while 3D CNNs employ the 3D convolutions to capture appearance and relation jointly and implicitly.

{\bf Models based on multiplicative interactions.} Modeling or learning correspondence is an important task in computer vision. Typically, these methods are fundamentally based on the multiplicative interactions between pixels or between filter responses~\cite{Memisevic13}. Mapping units~\cite{Hinton81a} first introduced the idea of multiplicative interactions to model relation between different views. Gated Boltzmann machines~\cite{MemisevicH07} were proposed to learn image transformation in unsupervised manner. Energy models~\cite{Energymodel}, which may be viewed as a way to emulate multiplicative interactions by computing squares, were proposed to model motion information in videos. Independent Subspace Analysis (ISA)~\cite{HyvarinenH00} was designed for invariant feature learning by computing sums over squared. ISA is similar to Energy model but its weights are trained from data. High-order neural networks~\cite{giles1987learning} were proposed to learn invariance based on polynomial expansions of input. Recently, some action recognition methods are based on energy models~\cite{DerpanisSCW10,WangQT13} and feature learning with Gate Boltzmann machines~\cite{TaylorFLB10} and ISA~\cite{LeZYN11}. Meanwhile, these multiplicative interactions or correlation models were integrated into the CNN architecture for optical flow estimation~\cite{DosovitskiyFIHH15} and person re-identification~\cite{LiZXW14}. 

Our proposed relation branch is inspired by these early works with multiplicative interactions and in particular it shares a similar {\em square-pooling} architecture with ISA. Our work differs from them in three important aspects: (1) The weights of relation branch are learned in a supervised manner with standard back propagation, while the previous work manually set model weights or learn them in an unsupervised manner. (2) The relation branch is integrated with an appearance branch to form the SMART block to capture spatiotemporal information, while previous works only has a module focusing on modeling relation. (3) We construct ARTNets by stacking multiple SMART blocks to learn hierarchical spatiotemporal features, while previous work usually has a single layer based on multiplicative interactions.

\section{Spatiotemporal Feature Learning}

In this section we describe our method for spatiotemporal feature learning. First, we discuss the role of multiplicative interaction in modeling relation across multiple frames. Next, we introduce the design of a SMART block. Finally, we propose the ART-Net by stacking multiple SMART blocks in the architecture of C3D-ResNet18.

\subsection{Multiplicative interactions}

\begin{figure*}
\includegraphics[width=\linewidth]{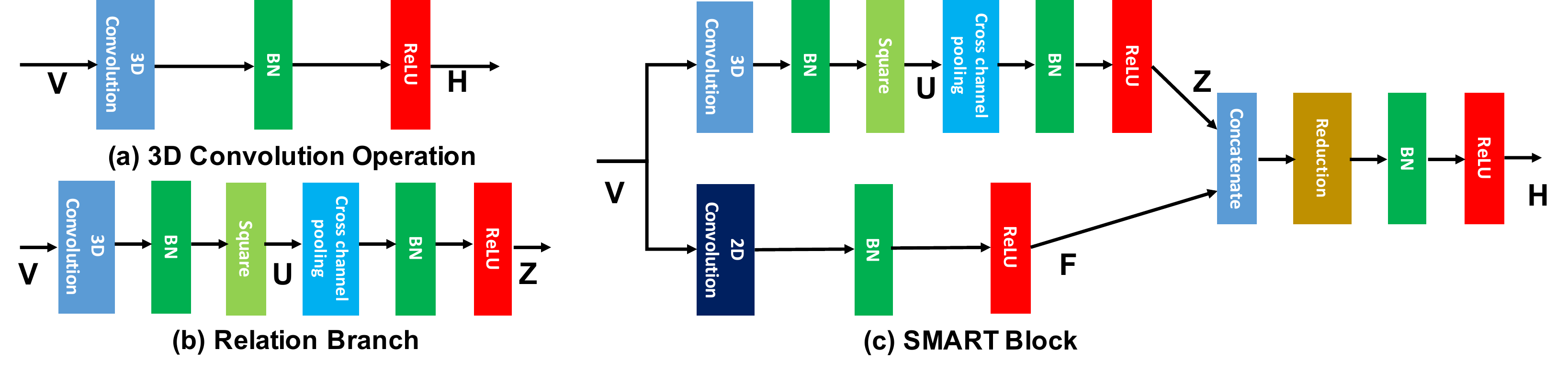}
\caption{{\bf Building blocks}: (a) the 3D convolution operation learns spatiotemporal features jointly and implicitly. (b) we first propose a {\em square-pooling} architecture to learn appearance-independent relation between frames. (c) we further construct a SMART block to learn spatiotemporal features separately and explicitly, where appearance branch uses a 2D convolution to capture static structure and relation branch employs a squaring-pooling architecture to model temporal relation.}
\label{fig:smart}
\vspace{-3mm}
\end{figure*}

Suppose we have two patches $\mathbf{x}$ and $\mathbf{y}$ from consecutive frames, we aim to learn the transformation (relation) $\mathbf{z}$ between them. A natural solution to this problem is to perform standard feature learning on the concatenation of these two patches, just like a 3D convolution~\cite{JiXYY13,TranBFTP15}, as follows:
\begin{equation}
z_k = \sum_{i} w_{ik}^x x_i + \sum_{j} w_{jk}^y y_j,
\end{equation}
where the transformation code $z_k$ is defined as a linear combination of $[\mathbf{x}, \mathbf{y}]$ by the parameters $\mathbf{w} = [\mathbf{w}^x_{k}, \mathbf{w}^y_{k}]$. However, in this case, the activation would be dependent on the appearance (content) of patches not just the transformation between them~\cite{Memisevic13}. In this sense, if both patches change but not transformation between them, the activation value would also change. Thus, this solution couples the information of appearance and relation together, adding the modeling difficulty and increasing the over-fitting risk.

Assuming the independence between appearance and relation, it is reasonable to decouple these two kinds of information when designing learning modules. It is easy to propose an appearance-independent relation detector by using {\em multiplicative interactions} between patches $\mathbf{x}$ and $\mathbf{y}$~\cite{Memisevic13}. Specifically, the transformation code $z_k$ could be defined as follows:
\begin{equation}
z_k = \sum_{ij} w_{ijk} x_i y_j,
\label{equ:mapping_unit}
\end{equation}
where the transformation code $z_k$ is defined as a linear combination of quadratic form from $\mathbf{x}$ and $\mathbf{y}$ by the weight tensor $w_{..k}$. Essentially, this transformation code $z_k$ pools over the outer product of $\mathbf{x}$ and $\mathbf{y}$, each element of which represents the evidence for a specific type of transformation. Thus, the activation value of $z_k$ is less likely dependent on the patch appearance than the transformation between them.  

{\bf Factorization and energy models.} The major obstacle to directly deploy Equation (\ref{equ:mapping_unit}) is that the number of parameters is roughly cubic in the number of pixels. Factorizing the parameter tensor $W$ into three matrices would be an efficient way to reduce model parameters~\cite{Memisevic13}, namely: $w_{ijk} = \sum_{f=1}^F w^x_{if} w^y_{jf} w^z_{kf}$. Thus, the transformation code $z_k$ between patch $\mathbf{x}$ and patch $\mathbf{y}$ in Equation (\ref{equ:mapping_unit}) would be rewritten as follows:
\begin{equation}
\begin{split}
z_k & = \sum_f w_{kf}^z \sum_i w_{if}^x x_i \sum_j w_{jf}^y y_j \\
& = \sum_f w_{kf}^z (\mathbf{w}^{x\mathrm{T}}_{f} \mathbf{x}) (\mathbf{w}^{y\mathrm{T}}_{f} \mathbf{y}).
\end{split}
\label{equ:factoration}
\end{equation}
This factorization formulation is closely related to energy model~\cite{Energymodel} and could be implemented with it. Specifically, a hidden unit $z_k$ in the energy model is calculated as follows:
\begin{displaymath}
\begin{split}
z_k = & \sum_f w^z_{kf} (\mathbf{w}^{x\mathrm{T}}_{f} \mathbf{x} + \mathbf{w}^{y\mathrm{T}}_{f} \mathbf{y})^2 \\
 = & \sum_{f} w_{kf}^z [2(\mathbf{w}^{x\mathrm{T}}_{f} \mathbf{x}) (\mathbf{w}^{y\mathrm{T}}_{f} \mathbf{y}) +   (\mathbf{w}^{x\mathrm{T}}_{f} \mathbf{x})^2 + (\mathbf{w}^{y\mathrm{T}}_{f} \mathbf{y})^2],
\end{split}
\end{displaymath}
where hidden units are the same with Equation (\ref{equ:factoration}) except the quadratic terms $(\mathbf{w}^{x\mathrm{T}}_{f} \mathbf{x})^2$ and $(\mathbf{w}^{y\mathrm{T}}_{f} \mathbf{y})^2$, which do not have a significant effect on the hidden unit~\cite{Memisevic12}. This energy model could be efficiently implemented with standard operations in 3D CNNs and easily stacked layer-by-layer as introduced in the next subsections.

\subsection{SMART blocks}

As shown in Figure~\ref{fig:smart}, a SMART block is a basic computational unit operating on an input volume $\mathbf{V} \in \mathbb{R}^{W \times H \times T \times C}$ and producing an output volume $\mathbf{H} \in \mathbb{R}^{W' \times H' \times T' \times C'}$. The motivation of the SMART block is to simultaneously model appearance and relation in a separate and explicit manner. Specifically, it learns spatiotemporal features from volume input with a two-branch architecture: (1) appearance branch for spatial feature learning, and (2) relation branch for temporal feature learning. 

{\bf Appearance branch} operates on individual frames and aims at capturing useful static information for action recognition. Static cues are sometimes important for action recognition as some action classes are strongly associated with certain object or scene categories. Specifically, we employ a {\em 2D convolution} to operate on the video volume $\mathbf{V}$ to capture the spatial structure in each frame. The output of 2D convolution is a volume $\mathbf{F} \in \mathbb{R}^{W_s \times H_s \times T_s \times C_s}$. The response values $\mathbf{F}$ of 2D convolution usually goes into another Batch Normalization (BN)~\cite{IoffeS15} layer and Rectified Linear Unit (ReLU)~\cite{NairH10} for non-linearity.

{\bf Relation branch} operates on stacked consecutive frames and aims to capture the relation among these frames for action recognition. This relation (transformation) information is crucial for action understanding as it contains motion cues. According to the discussion on multiplication interactions in the previous subsection, we design a {\em square-pooling} architecture to model temporal relation on this volume input. Specifically, we first apply a {\bf 3D convolution} to this volume input $\mathbf{V}$, which further goes through a {\bf square function} to obtain hidden units $\mathbf{U} \in \mathbb{R}^{W_t \times H_t \times T_t \times C_t}$. Then, we apply a {\bf cross-channel pooling}  to aggregate multiple hidden units in $\mathbf{U}$ into the transformation codes $\mathbf{Z} \in \mathbb{R}^{W_t \times H_t \times T_t \times C'_t}$. This cross-channel is implemented with a $1 \times 1 \times 1$ convolution. In practice, the transformation code $\mathbf{Z}$ would also go through a BN layer and ReLU non-linearity to be consistent with the output of appearance branch. Meanwhile, we also add a BN layer between the 3D convolution and the square non-linearity to improve its stability.

\begin{table*}
\centering
\resizebox{0.9\textwidth}{!}{
\begin{tabular}{c|c|c|c|c}
\hline
layer name & output size & C3D-ResNet18 & ARTNet-ResNet18 (s) & ARTNet-ResNet18 (d) \\ 
\hline
conv1 & $56 \times 56 \times 8$ & 3D conv $ 7 \times 7 \times 3 $, stride $2 \times 2 \times 2$ & \multicolumn{2}{c}{SMART $7 \times 7 \times 3$, stride $2 \times 2 \times 2$} \\
\hline
&  & & & \\
conv2\_x &   $ 56 \times 56 \times 8$ &  $ \begin{bmatrix} \mathrm{3D \ conv} & 3 \times 3 \times 3 & 64 \\  \mathrm{3D \ conv} & 3 \times 3 \times 3 & 64 \end{bmatrix} \times 2 $  & $\begin{bmatrix} \mathrm{3D \ conv} & 3 \times 3 \times 3 & 64 \\  \mathrm{3D \ conv} & 3 \times 3 \times 3 & 64 \end{bmatrix} \times 2 $ & $\begin{bmatrix} \mathrm{3D \ conv} & 3 \times 3 \times 3 & 64 \\  \mathrm{SMART} & 3 \times 3 \times 3 & 64 \end{bmatrix} \times 2 $ \\
&  & & & \\
\hline
&  & & & \\
conv3\_x & $ 28 \times 28 \times 4$ & $\begin{bmatrix} \mathrm{3D \ conv} & 3 \times 3 \times 3 & 128 \\  \mathrm{3D \ conv} & 3 \times 3 \times 3 & 128 \end{bmatrix} \times 2 $  & $\begin{bmatrix} \mathrm{3D \ conv} & 3 \times 3 \times 3 & 128 \\  \mathrm{3D \ conv} & 3 \times 3 \times 3 & 128 \end{bmatrix} \times 2 $ & $\begin{bmatrix} \mathrm{3D \ conv} & 3 \times 3 \times 3 & 128 \\  \mathrm{SMART} & 3 \times 3 \times 3 & 128 \end{bmatrix} \times 2 $ \\
&  & & & \\
\hline
&  & & & \\
conv4\_x & $ 14 \times 14 \times 2$ & $ \begin{bmatrix} \mathrm{3D \ conv} & 3 \times 3 \times 3 & 256 \\  \mathrm{3D \ conv} & 3 \times 3 \times 3 & 256 \end{bmatrix} \times 2 $  & $\begin{bmatrix} \mathrm{3D \ conv} & 3 \times 3 \times 3 & 256 \\  \mathrm{3D \ conv} & 3 \times 3 \times 3 & 256 \end{bmatrix} \times 2 $ & $\begin{bmatrix} \mathrm{3D \ conv} & 3 \times 3 \times 3 & 256 \\  \mathrm{SMART} & 3 \times 3 \times 3 & 256 \end{bmatrix} \times 2 $ \\
&  & & & \\
\hline
&  & & & \\
conv5\_x & $ 7 \times 7 \times 1$ & $\begin{bmatrix} \mathrm{3D \ conv} & 3 \times 3 \times 3 & 512 \\  \mathrm{3D \ conv} & 3 \times 3 \times 3 & 512 \end{bmatrix} \times 2 $  & $\begin{bmatrix} \mathrm{3D \ conv} & 3 \times 3 \times 3 & 512 \\  \mathrm{3D \ conv} & 3 \times 3 \times 3 & 512 \end{bmatrix} \times 2 $ & $\begin{bmatrix} \mathrm{3D \ conv} & 3 \times 3 \times 3 & 512 \\  \mathrm{3D \ conv} & 3 \times 3 \times 3 & 512 \end{bmatrix} \times 2 $ \\
&  & & & \\
\hline
 & $1 \times 1 \times 1$ & \multicolumn{3}{c}{average pool, dropout, 400-d fc, softmax} \\
\hline \hline
params (M) &  - & 33.37 & 33.39 & 35.20 \\ \hline
FLOPs (G) & - & 19.58 & 19.97 & 23.70  \\
\hline
\end{tabular}
}
\vspace{1mm}
\caption{{\bf Architectures for the Kinetics dataset}: we study three different networks for spatiotemporal feature learning by stacking two types of building blocks from Figure~\ref{fig:smart}: (1) 3D convolutions and (2) SMART blocks. Building blocks are shown in brackets, with the numbers of stacked blocks. The input to these networks is volume of $112 \times 112 \times 16$ and downsample is performed conv3\_1, conv4\_1, and conv5\_1 with a stride of $2 \times 2 \times 2$. }
\label{tbl:artnet-resnet18}
\vspace{-3mm}
\end{table*}

{\bf A SMART block} combine the output of the appearance and relation branches with a concatenation and reduction operation. Intuitively, the spatial and temporal features are complementary for action recognition and this fusion step aims to compress them into a more compact representation. In particular, we employ a $1 \times 1 \times 1 $ convolution on the concatenation volume $[\mathbf{F}, \mathbf{Z}] \in \mathbb{R}^{W' \times H' \times T' \times (C_s+C'_t)}$ to obtain the compressed feature volumes $\mathbf{H} \in \mathbb{R}^{W' \times H' \times T' \times C_f}$. As a common practice, this compressed feature volume $\mathbf{H}$ further goes through a BN layer and ReLU activation function.

{\bf Implementation details.} For the design simplicity of SMART block, some default setting is fixed as follows. First, the spatial and temporal dimension of output of two branches are ensured to be the same for concatenation operation, {\em i.e.,} $W_s=W_t=W'$, $H_s=H_t=H'$, and $T_s = T_t=T'$. In this sense, we let stride of 2D convolution in appearance branch and 3D convolution in relation branch be the same. Meanwhile, the spatial dimension of convolution kernels from two branches are the same as well. Second, the number of 2D convolution kernels in appearance branch is the same with that of 3D convolution kernels in relation branch, {\em i.e.,} $C_s = C_t$. In cross-channel pooling layer, each transformation code $z_k$ in relation branch is locally connected a group of hidden units instead of using full connectivity. The number of transformation code is set to be half of that of hidden unit $u$, {\em i.e.,} $C_t = 2C'_t$, and thereby the group size is set to be $2$. The weights in cross-channel pooling are fixed as $0.5$. Finally, for the output of SMART block, we set its output number to be equal to that of appearance branch, {\em i.e.,} $C_f = C_s$. Therefore, the design parameter of a SMART block is the same with that a normal 3D convolution, including kernel size $k \times k \times t$, convolutional stride $s_s$ and $s_t$, the output number $c$.

\subsection{Exemplars: ARTNet-ResNet18}

After introducing the SMART block, we are ready to plug it into the existing network architecture to build the appearance-and-relation network (ARTNet). The flexibility of the SMART block allows it to replace the role of a 3D convolution in learning spatiotemporal feature. In current implementation, we develop an ARTNet by integrating the SMART block into the C3D-ResNet18 architecture~\cite{Tran17}, and thereby the resulted architecture is coined as {\bf ARTNet-ResNet18}. 

We choose the C3D-ResNet18 to instantiate the ARTNet and the architecture details are shown Table~\ref{tbl:artnet-resnet18}. These networks take an $112 \times 112 \times 16$ input to keep a balance between model capacity and processing efficiency.
To well evaluate the effectiveness of SMART block, we implement two kinds of ARTNet-ResNet18: (1) we only replace the first 3D convolution in C3D-ResNet18 with the SMART block while keep the remaining layers unchanged, denoted as {\bf ARTNet-ResNet18 (s)}. (2) we stack multiple SMART blocks and totally replace seven 3D convolutions, denoted as {\bf ARTNet-ResNet18 (d)}. Stacking multiple SMART blocks allows us to capture appearance and relation information from different scales and further enhance the modeling capacity of ARTNet-ResNet18 (s).

{\bf Implementation details.} We test these networks on the recently introduced Kinetics dataset~\cite{KayCSZHVVGBNSZ17}. {\em All these models are trained on the train set of Kinetics dataset from scratch}. We train the C3D-ResNet18 and ARTNet-ResNet18 by following the common practice in~\cite{Tran17,KayCSZHVVGBNSZ17}. The network parameters are initialized randomly. We use the mini-batch stochastic gradient descent algorithm to learn network parameters, where the batch size is set to $256$ and momentum is set to $0.9$. The frames are resized to $128 \times 170$ and then a volume of $112 \times 112 \times 16$ is randomly trimmed and cropped from each training video. This volume also undergoes a random horizontal flip, with the per-pixel mean subtracted. The learning rate is initialized as $0.1$ and divided by a factor of $10$ when validation loss saturates. The total number of iteration is $250,000$ on the Kinetics dataset. To reduce the risk of over-fitting, we add a dropout layer before the final classification layer, where the dropout ratio is set to $0.2$. 

For testing network, we follow the common evaluation scheme~\cite{Tran17,SimonyanZ14}, where we sample $250$ volumes of $112 \times 112 \times 16$ from the whole video. Specifically, we first uniformly trim $25$ clips of $128 \times 170 \times 16$ and then generate $10$ crops of $112 \times 112 \times 16$ from each clip (4 corners, 1 center, and their horizontal flipping). The final prediction result is obtained by taking an average over these $250$ volumes.

\section{Experiments}

In this section we describe the experimental results of our method. First, we introduce the action recognition datasets and the evaluation settings. Then, we study different aspects of our proposed ARTNets on the Kinetics dataset and compare with the state-of-the-art methods. Finally, we transfer the learned spatiotemporal representations in ARTNets to the datasets of UCF101 and HMDB51.

\subsection{Datasets}

We evaluate the performance of ARTNets on three action recognition benchmarks: (1) Kinetics~\cite{KayCSZHVVGBNSZ17}, (2) UCF101~\cite{Soomro12}, and (3) HMDB51~\cite{KuehneJGPS11}. The Kinetics dataset is the largest well-labeled action recognition dataset. Its current version contains $400$ action classes and each category has at least $400$ videos. In total, there are around $240,000$ training videos, $20,000$ validation videos, and $40,000$ testing videos. The evaluation metric on the Kinetics dataset is the average of top-1 and top-5 error. As Kinetics is the largest available dataset, we mainly study different aspects of ARTNets on this dataset {\bf with only RGB input under the setting of training from scratch}.

UCF101 and HMDB51 are another two popular action recognition datasets, whose sizes are relatively small and the performance on them is already very high. The UCF101 has $101$ action classes and $13, 320$ video clips. We follow the official evaluation scheme and report average accuracy over three training/testing splits. The HMDB51 dataset is a collection of realistic videos from various sources, including movies and web videos. This dataset has $6,766$ videos from $51$ action categories. Our experiment follows the original evaluation scheme using three training/testing splits and reports the average accuracy. As these two datasets are relatively small, we cannot train ARTNets from scratch and thereby transfer the video representations learned from the Kinetics dataset to them by fine tuning. The fine-tuning process follows the good practice presented in the temporal segment networks (TSN)~\cite{WangXWQLTV16}. The goal of experiment on UCF101 and HMDB51 is to test the generalization ability of learned spatiotemporal features by the ARTNet. 

\subsection{Results on the Kinetics dataset}

\begin{table}
\centering
\resizebox{0.4\textwidth}{!}{
  \begin{tabular}{|l|c|c|c|}
    \hline
    Method &  Top-1 &  Top-5 & Avg \\
    \hline
    \hline
    C2D-ResNet18 &  61.2\% & 82.6\% & 71.9\% \\
    \hline
    C3D-ResNet18 & 65.6\% & 85.7\% & 75.7\% \\
    \hline
    C3D-ResNet34 & 67.1\% & 86.9\% & 77.0\% \\
    \hline
    \hline
    Relation-ResNet18 (s) & 67.5\% & 87.0\% & 77.2\% \\
    \hline
    Relation-ResNet18 (d) & 67.1\% & 86.7\% & 76.9\% \\
    \hline
    \hline
    ARTNet-ResNet18 (s)  & 67.7\% & 87.1\% & 77.4\% \\
    \hline
    ARTNet-ResNet18 (d)  & {\bf 69.2\%} & {\bf 88.3\%} & {\bf 78.7\%} \\
    \hline
  \end{tabular}
}
\vspace{1mm}
\caption{Comparison of ARTNet and C3D on the validation set of Kinetics dataset. We investigate the performance of basic blocks, including: 2D convolution, 3D convolution, relation branch, and SMART. We also study the effect of the stacking depth of the ARTNet. The performance is measured by {\bf Top-1 and Top-5 accuracy}.}
\label{tbl:block}
\vspace{-3mm}
\end{table}

{\bf Study on building block.} We begin our experiment by studying the performance of four building blocks for spatiotemporal feature learning in videos. These building blocks include: (1) 2D convolution, (2) 3D convolution, (3) Relation branch, and (4) SMART block. We conduct experiments on Kinetics with the ResNet18 architecture as shown in Table~\ref{tbl:artnet-resnet18}. For C2D-ResNet18, we replace the 3D convolutions in C3D-ResNet18 with 2D convolutions, and for Relation-ResNet18, we replace the SMART blocks in ARTNet-ResNet18 with relation branch. The results are shown in Table~\ref{tbl:block}. We see that 3D convolutions outperforms 2D convolutions for learning video representations ($75.7\%$ vs. $71.9\%$). Our newly designed relation branch and SMART block both outperform the original 3D convolutions ($77.2\%$ vs. $75.7\%$ and $77.4\%$ vs. $75.7\%$). SMART block obtains the best performance among these four building blocks, demonstrating the effectiveness of modeling appearance and relation separately and explicitly.

{\bf Study on block stacking.} We also investigate the effectiveness of stacking multiple Relation branches and SMART blocks. As shown in Table~\ref{tbl:block}, we observe that stacking multiple SMART blocks is able to further boost error rate from $77.4\%$ to $78.7\%$. This improvement indicates the effectiveness of capturing spatiotemporal features in a hierarchical manner. However, stacking multiple relation branch causes a small performance drop, indicating the importance of modeling spatial structure in higher layers. Remarkably, as stacking SMART blocks would increase the network depth, we also compare the performance with C3D-ResNet34 in Table~\ref{tbl:block}, where ARTNet-ResNet18 even outperforms the deeper C3D-ResNet34 ($78.7\%$ vs. $77.0\%$). This result demonstrates that the performance improvement is brought by the effectiveness of SMART block instead of the increased network depth. In the remaining experiments, we will use the ARTNet-ResNet18 (d) by default.

\begin{table}
\centering
\resizebox{0.45\textwidth}{!}{
  \begin{tabular}{|l|c|c|c|c|}
    \hline
    Method &  Modality & Top-1 &  Top-5 & Avg \\
    \hline
    \hline
    C3D-ResNet18 & RGB & 65.6\% & 85.7\% & 75.7\% \\
    \hline
    C3D-ResNet18 & Flow & 57.5\% & 80.6\% & 69.0\% \\
    \hline
    C3D-ResNet18 & Fusion & 68.7\% & 87.8\% & 78.2\% \\
    \hline
    \hline
    ARTNet-ResNet18 & RGB & 69.2\% & 88.3\% & 78.7\% \\
    \hline
    ARTNet-ResNet18 & Flow & 59.8\% & 82.3\% & 71.0\% \\
    \hline
    ARTNet-ResNet18 & Fusion & {\bf 71.3\%} & {\bf 89.5\%} & {\bf 80.4\%} \\
    \hline
  \end{tabular}
}
\vspace{1mm}
\caption{Comparison of ARTNet and C3D with two stream input, i.e., RGB and Optical Flow. The results are reported on the validation set of Kinetics with the measure of {\bf Top-1 and Top-5 accuracy}.}
\label{tbl:two-stream}
\vspace{-3mm}
\end{table}

{\bf Study on two-stream inputs.} Two stream CNN is a strong baseline for action recognition and its input has two modalities, {\em i.e.}, RGB and Optical Flow. To further illustrate the effectiveness of SMART block over 3D convolution, we perform experiments with two-stream inputs for both ARTNet-ResNet18 and C3D-ResNet18. The numerical results are reported in Table~\ref{tbl:two-stream}. {\em First}, we find that two-stream inputs are able to improve the performance of C3D-ResNet18 from $75.7\%$ to $78.2\%$. This improvement indicates that although 3D convolution aims to directly learn spatiotemporal features from RGB, flow stream is still able to provide complementary information. {\em Second}, comparing two-stream C3D-ResNet18 with RGB-stream ARTNet-ResNet18, we notice that our proposed ARTNet is still able to yield a slightly better performance ($78.7\%$ vs. $78.2\%$). This better result demonstrates the superiority of SMART block over two stream inputs. {\em Finally}, we also experiment ARTNet-ResNet18 with two-stream inputs. In flow stream, similar improvement over C3D-ResNet18 is also observed with ARTNet-ResNet18. The two-stream ARTNet-ResNet18 can boost performance to $80.4\%$. But it is worth noting that the high computational cost of optical flow makes it extremely difficult to apply at large-scale datasets and deploy in real-world applications. Therefore, in the remaining experiment, we mainly compare the performance of only using RGB input.

\begin{table}
\centering
\resizebox{0.48\textwidth}{!}{
  \begin{tabular}{|l|c|c|c|c|c|}
    \hline
    Method &  TSN & Modality & Top-1 &  Top-5 & Avg \\
    \hline
    ARTNet-ResNet18 & No & RGB & 69.2\% & 88.3\% & 78.7\% \\
    \hline
    ARTNet-ResNet18 & Yes & RGB & 70.7\% & 89.3\% & 80.0\% \\
    \hline
    \hline
    ARTNet-ResNet18 & No & Flow & 59.8\% & 82.3\% & 71.0\% \\
    \hline
    ARTNet-ResNet18 & Yes & Flow & 60.6\% & 83.1\% & 71.9\% \\
    \hline
     \hline
    ARTNet-ResNet18 & No & Fusion & 71.3\% & 89.5\% & 80.4\% \\
    \hline
    ARTNet-ResNet18 & Yes & Fusion & {\bf 72.4\%} & {\bf 90.4\%} & {\bf 81.4\%} \\
    \hline
  \end{tabular}
}
\vspace{1mm}
\caption{Comparison of ARTNet between without TSN and with TSN. ARTNet focuses on short-term spatiotemporal feature learning and is easily combined with the existing long-term modeling architectures. The results are reported on the validation set of Kinetics and measured by {\bf Top-1 and Top-5 accuracy}.}
\vspace{-3mm}
\label{tbl:tsn}
\end{table}

{\bf Study on long-term modeling.} The proposed SMART block and ARTNet focus on short-term spatiotemporal feature learning and is complementary to the exiting long-term modeling architectures~\cite{VarolLS16,WangXWQLTV16,NgHVVMT15}. Temporal segment network (TSN) is a general and flexible video-level framework for learning action models in videos~\cite{WangXWQLTV16}. The simplicity nature of TSN allows us to replace the original 2D CNNs with our proposed ARTNet-ResNet18. Specifically, to keep a balance between modeling capacity and training time, we set the segment number as $2$. The experimental results are summarized in Table~\ref{tbl:tsn}. We see that TSN modeling is helpful to improve the performance of ARTNet-ResNet18. For example, ARTNet-ResNet18 with TSN training can yield the performance of $80.0\%$ with RGB input and $81.4\%$ with two-stream inputs, which is better than the original performance by $1.3\%$ and $1.0\%$. This improvement demonstrates the complementarity of ARTNet to the TSN framework. In principle, ARTNet is a general short-term video model, that could be explored in any long-term learning framework, such as LSTM~\cite{NgHVVMT15,DonahueHGRVDS15} and attention modeling~\cite{WangXLV17}.

{\bf Comparison to the state of the art.} We compare the performance of ARTNet-ResNet18 with the state-of-the-art approaches on the validation set and test set of Kinetics. The results are summarized in Table~\ref{tbl:sota1}. For fair comparison, we consider methods that only use RGB input and learned from scratch on the train set of Kinetics. We also list other important factors such as spatial resolution and backbone architectures. 

We {\em first} compare with three baseline methods: (1) CNN+LSTM~\cite{NgHVVMT15,DonahueHGRVDS15}, (2) Spatial Stream~\cite{SimonyanZ14}, and (3) C3D~\cite{TranBFTP15}. Our proposed ARTNets significantly outperform these baselines by around $10\%$. We {\em then} compare with deeper C3D architecture~\cite{Tran17} such as C3D-ResNet18 and C3D-ResNet34. Our ARTNet is able to yield a better performance (around $3\%$) than these fairly-comparable models. {\em Finally}, we compare with the recent state-of-the-art methods, namely temporal segment network (TSN)~\cite{WangXWQLTV16} and Inflated 3D CNN (I3D)~\cite{CarreiraZ17}. These two architectures employ a deeper backbone architecture (Inception~\cite{SzegedyLJSRAEVR15}) and larger spatial resolution ($224 \times 224$). Besides, I3D is also equipped with long-term modeling~\cite{VarolLS16} by stacking $64$ frames. Therefore, it is fair for us to use TSN to increase the temporal duration of ARTNet. Our ARTNet with TSN training obtains a slightly better performance than these two very competitive methods ($80.0\%$ vs. $77.8\%$ on validation set, and $78.7\%$ vs. $78.2\%$ on test set). 

It is worth noting that the current published state-of-the-art performance is $82.7\%$, that is obtained by two-stream I3D~\cite{CarreiraZ17} with optical flow input and pre-training on ImageNet. Two-stream I3D is more computational expensive than ARTNet as it uses larger spatial resolution, deeper structure, and two-stream inputs. The winner solution~\cite{winner17} at ActivityNet challenge~\cite{ghanem2017activitynet} gets a performance of $87.6\%$ by using more modalities, multi-stage training, and model ensemble. These results are not directly comparable to ours. 

\begin{table*}
\resizebox{\textwidth}{!}{
  \begin{tabular}{|l|c|c|c|c|}
    \hline
    Method & Spatial resolution & Backbone architecture & Kinetics val set & Kinetics test set \\
    \hline
    \hline
    ConvNet+LSTM~\cite{DonahueHGRVDS15,NgHVVMT15} & $299 \times 299$ & ResNet-50 & - & 68.0\% \\ 
    \hline
    Two Stream Spatial Networks~\cite{SimonyanZ14} &  $299 \times 299$ & ResNet-50 & - & 66.6\% \\
    \hline
    C3D~\cite{TranBFTP15} & $112 \times 112$ & VGGNet-11 & - & 67.8\% \\
    \hline
    C3D~\cite{Tran17}  & $112 \times 112$ & ResNet-18 & 75.7\% & 74.4\% \\
    \hline
    C3D~\cite{Tran17}  & $112 \times 112$ & ResNet-34 & 77.0\% & 75.3\% \\
    \hline
    TSN Spatial Networks~\cite{WangXWQLTV16} & $224 \times 224$ & Inception V2 & 77.8\% & - \\
    \hline
    RGB-I3D~\cite{CarreiraZ17} & $224 \times 224$ & Inception V1 & - & 78.2\% \\
    \hline
    \hline
    ARTNet w/o TSN & $112 \times 112$ & ResNet-18 &  78.7\% & 77.3\% \\
    \hline
    ARTNet with TSN & $112 \times 112$ & ResNet-18 & {\bf 80.0\%} & {\bf 78.7\%}  \\
    \hline
  \end{tabular}
}
\vspace{1mm}
\caption{Comparison with state-of-the-art methods on the validation and test set of Kinetics. The performance is measured by the average of Top-1 and Top-5 accuracy. {\bf For fair comparison, we consider methods that use only RGB input and train from scratch on Kinetics}. Our ARTNets are trained from the spatial resolution of $112 \times 112$ and can still yield better performance than those trained from the spatial resolution of $224 \times 224$ or $229 \times 229$. }
\label{tbl:sota1}
\vspace{-3mm}
\end{table*}

\begin{table*}
\resizebox{\textwidth}{!}{
  \begin{tabular}{|l|c|c|c|c|c|}
    \hline
    Method & Pre-train dataset & Spatial resolution & Backbone architecture &  ~~UCF101~~ & HMDB51\\
    \hline
    \hline
    HOG~\cite{WangS13a} & None & $240 \times 320$ & None & 72.4\% & 40.2\% \\
    \hline
    \hline
    ConvNet+LSTM~\cite{DonahueHGRVDS15} & ImageNet & $224 \times 224$ & AlexNet & 68.2\%  & -  \\ 
    \hline
    Two Stream Spatial Network~\cite{SimonyanZ14} & ImageNet & $224 \times 224$ & VGG-M & 73.0\% & 40.5\% \\
    \hline
    Conv Pooling Spatial Network~\cite{FeichtenhoferPZ16} & ImageNet & $224 \times 224 $& VGGNet-16 & 82.6\% & - \\
    \hline
    Spatial Stream ResNet~\cite{FeichtenhoferPW16} & ImageNet & $224 \times 224$ & ResNet-50 & 82.3\% & 43.4\% \\
    \hline
    Spatial TDD~\cite{Wang0T15} & ImageNet & $224 \times 224$& VGG-M & 82.8\% & 50.0\% \\
    \hline
    RGB-I3D~\cite{CarreiraZ17} & ImageNet&$224 \times 224$ & Inception V1 &  84.5\% &  49.8\%  \\
    \hline
    TSN Spatial Network~\cite{WangXWQLTV16} & ImageNet & $224 \times 224$ & Inception V2 & {\bf 86.4\%} & {\bf 53.7\%} \\
    \hline
    \hline
    Slow Fusion~\cite{KarpathyTSLSF14} & Sports-1M & $ 170 \times 170 $ & AlexNet & 65.4\% & - \\
    \hline
    C3D~\cite{TranBFTP15} & Sports-1M & $112 \times 112$ & VGGNet-11 & 82.3\% & 51.6\% \\
    \hline
    LTC~\cite{VarolLS16} & Sports-1M & $71 \times 71$ & VGGNet-11 & 82.4\% & 48.7\% \\
    \hline
    C3D~\cite{Tran17} & Sports-1M & $112 \times 112$ & ResNet-18 & {\bf 85.8\%} & {\bf 54.9\%} \\
    \hline
    \hline
    TSN Spatial Network~\cite{WangXWQLTV16} & ImageNet+Kinetics & $224 \times 224$ & Inception V2 & 91.1\% & - \\ 
    \hline
    TSN Spatial Network~\cite{WangXWQLTV16} & ImageNet+Kinetics & $229 \times 229$ & Inception V3 & 93.2\% & - \\ 
    \hline
    RGB-I3D~\cite{CarreiraZ17} & ImageNet+Kinetics &$224 \times 224$ & Inception V1 & {\bf 95.6\%} & {\bf 74.8\%}  \\
    \hline
    \hline
    C3D & Kinetics & $112 \times 112$ & ResNet-18 & 89.8\% & 62.1\% \\
    \hline
    ARTNet w/o TSN & Kinetics & $112 \times 112$ & ResNet-18 & 93.5\% & 67.6\% \\
    \hline
    ARTNet with TSN & Kinetics & $112 \times 112$ & ResNet-18 & {\bf 94.3\%} & {\bf 70.9\%} \\
    \hline
  \end{tabular}
}
\vspace{1mm}
\caption{Comparison with state-of-the-art methods on the UCF101 and HMDB51 datasets. The accuracy is reported as average over three splits. {\bf For fair comparison, we consider methods that use only RGB input and pre-train on different datasets}. The performance is grouped according to its pre-training dataset. Our ARTNet obtains the best performance under the setting of pre-training only on the Kinetics dataset, and a comparable performance to the RGB-I3D pre-trained on the datasets of ImageNet+Kinetics.}
\label{tbl:sota2}
\vspace{-4mm}
\end{table*}

\subsection{Results on the UCF101 and HMDB51 datasets}

In this subsection we study the generalization ability of learned spatiotemporal representations on the Kinetics dataset~\cite{KayCSZHVVGBNSZ17}. Specifically, we transfer the learned models to two popular action recognition benchmarks: UCF101~\cite{Soomro12} and HMDB51~\cite{KuehneJGPS11}. We consider fine tuning three models trained on the Kinetics dataset: C3D-ResNet18, ARTNet-ResNet18 without TSN, ARTNet-ResNet18 with TSN. The fine-tuning process is conducted with the TSN framework and follows the common practice proposed in the original TSN framework~\cite{WangXWQLTV16}, where the segment number is set to 2.

The results are summarized in Table~\ref{tbl:sota2}. {\em First}, we compare the performance of C3D-ResNet18 and ARTNet-ResNet18 pre-trained on the Kinetics dataset and see that our ARTNet outperform C3D by $3.7\%$ on the UCF101 dataset and by $5.5\%$ on the HMDB51 dataset. This superior performance demonstrates that the spatiotemporal representation learned in ARTNet is more effective than C3D for transfer learning. {\em Then}, we investigate the ARTNet-ResNet18 models learned under the TSN framework on the Kinetics dataset and these models can yield a slightly better performance ($94.3\%$ on UCF101 and $70.9\%$ on HMDB51). This better transfer learning performance on UCF101 and HMDB51 agrees with the original performance improvement on the Kinetics dataset as shown Table~\ref{tbl:sota1}, which indicates the importance of long-term modeling. {\em Finally}, we compare with other state-of-the-art methods that pre-train on different datasets. We see that the methods pre-trained on the Kinetics dataset significantly outperform those pre-trained on  ImageNet~\cite{DengDSLL009} and Sports-1M~\cite{KarpathyTSLSF14}, which might be explained by the better quality of Kinetics. Our ARTNet obtains a comparable performance to the best performer of RGB-3D that is trained at a larger spatial resolution and pre-trained on two datasets (ImageNet and Kinetics).

\section{Conclusion and Future Work}
\vspace{-1mm}

In this paper we have presented a new architecture, coined as {\em ARTNet}, for spatiotemporal feature learning in videos. The construction of ARTNet is based on a generic building block, termed as {\em SMART}, which aims to model appearance and relation separately and explicitly with a two-branch unit. As demonstrated on the Kinetics dataset, SMART block is able to yield better performance than the 3D convolution, and ARTNet with a single RGB input even outperforms the C3D with two-stream inputs. For representation transfer from Kinetics to datasets of UCF101 and HMDB51, ARTNet also achieves superior performance to the original C3D. This performance improvement may be ascribed to the fact that we separately model appearance and relation, by using the linear combination of filter responses in each frame and the multiplicative interactions between filter responses across frames, respectively.

For ARTNet, augmenting RGB input with optical flow also helps to improve performance. This improvement indicates optical flow modality is still able to provide complementary information. However, the high computational cost of optical flow prohibits its application in real-world systems. In the future we plan to further improve the ARTNet architecture to overcome the performance gap between single-stream and two-stream inputs. Meanwhile, we will try to instantiate the ARTNets with more deeper structures such as ResNet101 and train them on more larger spatial resolutions.

{\small
\bibliographystyle{ieee}
\bibliography{egbib}
}

\end{document}